# Joint Semi-supervised RSS Dimensionality Reduction and Fingerprint Based Algorithm for Indoor Localization

C.F. ZHOU[1], L. MA[1,2], and X.Z. TAN[1,2]

[1]Harbin Institute of Technology Communication Research Center, HARBIN, 150000 CHN
[2]Key Laboratory of Police Wireless Digital Communication, Ministry of Public Security, HARBIN, 150000 CHN

## BIOGRAPHY (IES)

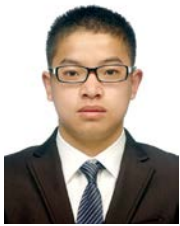

**C.F. ZHOU** received the B.S. degree in communication engineering from Harbin Institute of Technology (HIT), Harbin, China, in 2013. And now he is studying for his M.S. degree in HIT. His research interests include indoor position, feature extraction, pattern recognition and big data.

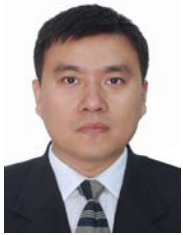

**L. Ma** is an associate professor in the Communication Research Center of Harbin Institute of Technology, China. He obtained his BSc, MSc, and PhD in Electrical Engineering from Harbin Institute of Technology, Harbin, China, in 2003, 2005, and 2009, respectively. Since 2013, he is a visiting scholar at the Edward S. Rogers Sr. Department of Electrical and Computer Engineering at the University of Toronto, Canada. His research interests include location-based service for wireless local area networks, cognitive radio, and cellular networks.

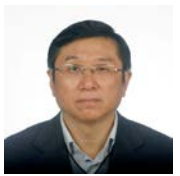

**X.Z. Tan (M'10)** received the B.S., M.S., and Ph.D. degrees from Harbin Institute of Technology, Harbin, China, in 1982, 1986, and 2005, respectively. From October 1988 to March 1990, he was a Visiting Scholar with Kyoto University, Kyoto, Japan. He is currently a Professor with the School of Electronics and Information Engineering, Harbin Institute of Technology. He is a Senior Member of the Chinese Institute of Electronics and the Chinese Institute of Communication. He had published more than 70 papers in international journals. His research interests include wireless communications, digital trunking communication, and cognitive radio.

## ABSTRACT

With the recent development in mobile computing devices and as the ubiquitous deployment of access points(APs) of Wireless Local Area Networks(WLANs), WLAN based indoor localization systems(WILSs) are of mounting concentration and are becoming more and more prevalent for they do not require additional infrastructure. As to the localization methods in WILSs, for the approaches used to localization in satellite based global position systems are difficult to achieve in indoor environments, fingerprint based localization algorithms(FLAs) are predominant in the RSS based schemes. However, the performance of FLAs has close relationship with the number of APs and the number of reference points(RPs) in WILSs, especially as the redundant deployment of APs and RPs in the system. There are two fatal problems, curse of dimensionality (CoD) and asymmetric matching(AM), caused by increasing number of APs and breaking down APs during online stage. In this paper, a semi-supervised RSS dimensionality reduction algorithm is proposed to solve these two dilemmas at the same time and there are numerous analyses about the theoretical realization of the proposed method. Another significant innovation of this paper is jointing the fingerprint based algorithm with CM-SDE algorithm to improve the localization accuracy of indoor localization. Comparing with LDE-KNN algorithm, SDE-KNN method is going to update RSS during online stage or offline stage. It is the update scheme that improves the performance of the proposed algorithm. After locally analyzing of parameters, the optimized value of each parameter could be gained. As it presents in by numeral analysis, the performance optimized parameters of SDE-KNN of it is overly better than initial KNN and LDE-KNN. The localization accuracy of SDE-KNN is higher than 70% as the error radius is 0.5 meter, and it is about 10 percent higher than initial KNN, as well as 25 percent higher than LDE-KNN. As the error radius equals to 1 meter, the proposed algorithm could gain over 95% of localization accuracy.

*Index Terms*—Dimensionality reduction; Fingerprint based localization algorithm; Indoor position; Kernel based fuzzy c-means (KFCM); Semi-supervised discriminant embedding (SDE); WLAN;

## I. INTRODUCTION

**W**ith the recent development in mobile computing devices and wireless techniques, indoor localizing systems are of mounting concentration and are becoming more and more prevalent[1]. The amalgamation of location information and advanced mobile computing devices facilitate the ubiquitous of location-based services (LBSs), such as personal safety, tourism, and content delivery[2~3]. Although satellites based global position systems, like global positioning system (GPS) have been in service for several decades, its indoor application has been limited by the unavailability of satellites in sight and poor positioning signals in indoor environments. Therefore, to complement the shortness of GPS in harsh indoor environment, several indoor positioning systems have been proposed in recent years, which are based on infrared[4], ultrasound[5~6], video[7~8], and radio frequency (RF)[9~10], especially indoor location systems based on wireless local area networks (WLANs). As the comprehensive applications of WLAN and redundant deployment of access points(APs), WLAN based indoor localization systems(WILPSs) attract numerous researchers and companies to pay attention on building its commercial systems[11] for they do not require additional infrastructure.

Approaches using to locate in indoor environments are similar to satellite based global position systems, for example, angle of arrival (AOA)[12], time of arrival (TOA)[13], and so on. However, received signal strength(RSS) based techniques are considered as predominant methods for WLAN indoor positioning, for the reason that the measurements of RSS, the realization of localization algorithms, and the deployment of systems are considerably easier than other approaches. RSS based indoor localization algorithm, also named as fingerprint based localization, can be generally classified into two categories: theoretical signal propagation model based approaches(TSPM)[14] and empirical model based methods (EM). Since there is no accurate model to estimate signal propagation in indoor environments, comparing to EMBM, TSPM has limited application in indoor localization[1].

Indoor localization systems based on empirical model always consist of two phases: offline and online phase. During offline phase, RSS from all reference points(RPs) is sampled in the interested area, and then all RP's coordinates and its corresponding RSS are saved as Radio Map, which is the fingerprint of RSS in the area. During online phase, RSS is tested by mobile terminals at test points(TPs) and then using fingerprint localization algorithms(FLAs) to match the RSS with Radio Map and it is similar to the process of pattern recognition. However, a dilemma comes as the increasing numbers of APs and RPs in the system. Taking $k$ nearest neighbors(KNN) fingerprint matching method as an example, its complexity has linear relationship with multiple of the number of APs and RPs. Thus, the time consumed at online stage will increase and it does meet the real time demands of indoor localization. Another drawback of the WILS comes out when several APs powered off or broken down during online stage. This is a fatal situation of the fingerprint matching algorithms for the RSS cannot match with Radio Map constructed during offline phase. These two weaknesses are called curse of dimensionality (CoD) and asymmetric matching(AM), respectively.

The solutions to the above stated problems emerged in little previous work. A simple way to solve AM problem is assuming that the RSS of broken AP equals to zero and then finishing the indoor localization. Although this approach could solve AM problem, it has miniature theoretical basis and cannot unravel CoD. In many fields like information processing, machine learning, data compression, scientific visualization, patter recognition and neural networks, the approaches to process hyper dimensional datasets, named dimensionality reduction(DR), are common[15], and it has two categories: linear and nonlinear. Principal component analysis(PCA)[16] and linear discriminant analysis(LDA)[17] belong to linear DR algorithms, however, local linear embedding(LLE)[18] and isometric mapping(ISOMAP)[19] are nonlinear DR methods. Comparing with linear DR algorithms, the advantages of nonlinear DR algorithms could preserve the nonlinear structures of hyper dimensional datasets.

However, most of DR algorithms, including above four typical ones, are usually used to visualize hyper datasets. A DR algorithm that could extract the features of hyper data sets is a sound choice to elucidate CoD. Chen el al proposed a supervised DR algorithm named local discriminant embedding(LDE)[20] which performed well in extracting features of face from high resolution photographs. And the authors of this paper applied LDE in the extraction the features of Radio Map and realized in reducing the complexity of online position as well as keeping the probability of localization almost the same[21]. While there is an obvious flaw that it needs high density sampled RSS to mine features of Radio Map efficiently since sampling massive RSS is a time consuming work, for example to sample a high density RSS data in the experiment region of this paper(See Section IV), it spent nearly one month(8 hours per day), up to 240 hours except the time of deploying the WILS.

In this paper, a semi-supervised discriminant embedding algorithm based on class matching(CM-SDE) is proposed in this paper, it could realize the function of extracting features under the condition that the sampling of RSS on RPs is sparse. It means that much shorter time is consumed if applying the proposed algorithm. The semi-supervise process of the proposed algorithm clearly shows the merits of CM-SDE, in other words, CM-SDE could refresh online localization database(OLD) according to the RSS which without class label, whatever RSS is sampled during online stage by mobile terminals or got at the offline stage by laptops. And then we joint CM-SDE with fingerprint based algorithm to locate the position of the user who are using indoor position services in the system. Comparing with LDE based Fingerprint algorithm(LDE-KNN) and the original fingerprint algorithm(KNN), as the simulating results reported, the performance of proposed algorithm(SDE-KNN) is overly better than KNN and LDE-KNN. The localization accuracy of SDE-KNN is higher than 70% as the error radius is 0.5 meter. And it is about 10 percent and 25 percent higher than initial KNN and LDE-KNN respectively. As the error radius equals to 1 meter, localization accuracy of the proposed algorithm is up to 95%.

The rest of this paper is as follows: Section II states the selected fingerprint based algorithm and its realization. In this section, the method to label the class of all RSS is also analyzed as well as its procedures. Section III presents the theoretical analysis of CM-SDE and its procedures. And it reports the flowchart of SDE algorithm applied in WLAN based localization systems. Section IV illustrates the numerical results of the simulation in the interested zone. In this section, the settings of experiments and the local optimization are analyzed before the analysis on the performance of proposed algorithm as well as LDE-KNN and KNN. Conclusions and future works are presented in Section V.

## II. BACKGROUND

### A. Fingerprint Based Localization Algorithm

Before stating FLAs, the author would like to briefly introduce about the process of building Radio Map. For the fingerprint based localization algorithm is a kind of supervised and matching approach, thereby there is not only RSS in the Radio Map, but also corresponding 2 dimensional coordinates of RSS. But it is not specified that FLAs just could apply in 2 dimension positioning, it could be generalized into 3 dimension situation easily. First step of constructing Radio Map is deploying APs and gridding RPs in the interested zone, and then coordinates of all RPs are marked and saved. As it shows in Fig.1, the density of grid is determined by the demand of localization accuracy in the hallway environment.

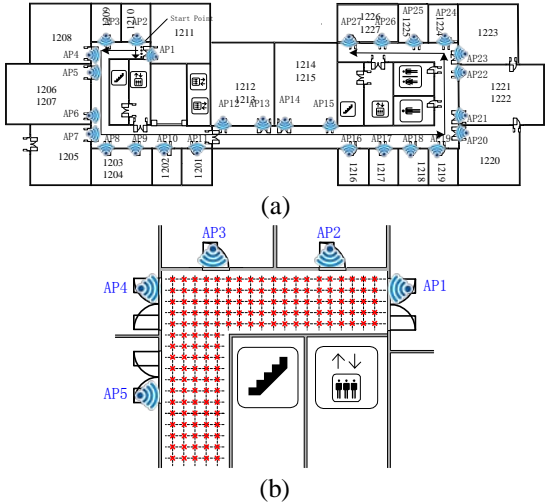

(a)

(b)

Fig.1: Deployment of WILS in the experimental zone and its sampling grid which interval equals 0.5 meter. (a): In the WLAN-based indoor, there are 27 APs and each AP works on 2.4GHz, which is Industrial Scientific Medical(ISM) band and is freedom of license. And the test zone is 12th floor of 2A building in scientific park of Harbin Institute of Technology (HIT). The interested area of indoor localization is the hallway. (b): By enlarging part of the test zone, the details of RPs, marked by red spot, are showed in it. In this paper, the interval of reference point is 0.5 meter.

And then a laptop or mobile ultimate with WiFi module is used to sample the RSS transmitted from all APs and then store them into specified format of file. The sampling process can be treated as a discrete time-sequence(after extracting the value of RSS) at the RP, its structure is shown in Fig.2. During the sampling stage, it might need to estimate the number of available RSS to ensure the efficiency of Radio Map.

Taking KNN as an example to fulfill the theory analysis of FLAs, it is the selected approach to compare with the propose algorithm. Procedures of KNN are as follows:

Step 1: receive RSS from all APs by the user and store as $RSS_{on}$;

Step 2: calculate Euclidean distance between $RSS_{on}$ and $N$ RPs' RSS in the hyperspace of RSS according to(1);

$$D_i(RSS_{on}, RSS_i) = \|RSS_{on} - RSS_i\|, i=1,\cdots,N \quad (1)$$

Step 3: find out $k$ RPs which has $k$ minimal Euclidean distance among all RPs;

Step 4: output the position of user $(x_U, y_U)$ by(2):

$$(x_U, y_U) = \sum_{i=1}^{k}(x_i, y_i)\Big/k \quad (2)$$

From above analysis of KNN, its complexity, by counting times of square, is $O(mN)$, where $m$ equals to the number of APs in the WILS.

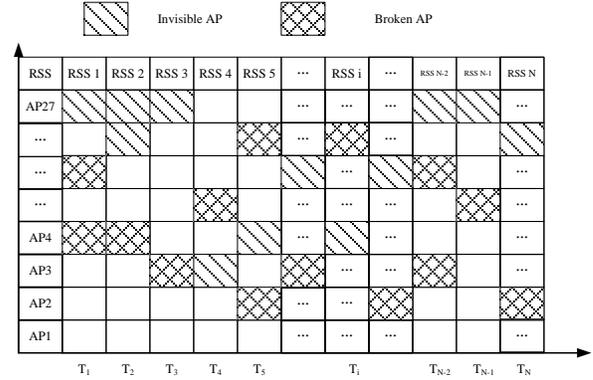

Fig.2: At a specified reference point, RSS is sampled by the mobile terminals or laptops. And the sampled RSS is treated as the discrete time-sequence. During the process of sampling RSS, there are several APs are invisible or broken down, especially during online stage for it could be ensure that all APs are opening and working on when constructing Radio Map.

### B. Classification by Clustering

As its stated in section I, the proposed method is on the basis of class matching. Therefore, an unsupervised classification algorithm, named kernel based fuzzy $c$-means (KFCM), is used to cluster RSS of RPs into specified number of clusters. Assuming that the RSS can be classified into $c$ clusters, the process of $c$-means cluster, which is the basis of KFCM, is carried out as follows:

Step 1: according to the format of RSS, generate $c$ centers of classes and termed as a vector $v = [v_1, v_2, \cdots, v_c]$;

Step 2: calculate the Euclidean distance between $RSS_i$, the RSS sampled at $ith$ RP, and all the centers by(1);

Step 3: labeling the $RSS_i$ by(3), where $G_i$ stands for the $ith$ class and $\arg\min\|\;\|$ means to find out minimal value of $\|\;\|$;

$$RSS_i \in G_i \Leftrightarrow \arg\min_{1\leq j\leq c} D(RSS_i, v_j) \quad (3)$$

Step 4: updating vector $v$ by(4), where $|G_i|$ means the number of RSS in $ith$ class;

$$v_i = \frac{1}{|G_i|}\sum_{RSS_k \in G_i} RSS_k \quad (4)$$

Step 5: iterate from step 2 to step 4 until the classification is convergent or the number of iterations is up to maximum times. The convergence conditions can be stated as(5), and $\varepsilon$ is a small positive number that limited the convergent radius:

$$\max \|v_i - v'_i\| \leq \varepsilon, \varepsilon \geq 0 \quad (5)$$

Above classification algorithm is a typical realization and it treats all RSS as equivalent, which means that the probability of RSS belongs to each class is the same. A fuzzy algorithm, however, treats all RSS as different, and the probability of RSS belongs to each class is dissimilar. In fuzzy $c$-means cluster, $U$ is a matrix that stands for the utility. And the process of calculating $U$ and updating vector $v$ is according to (6) where $m$ is nonlinear index:

$$\begin{cases} u_{ki} = \left[ \sum_{j=1}^{c} \left( \frac{D(RSS_k, v_i)}{D(RSS_k, v_j)} \right)^{1/m-1} \right]^{-1} \\ v_i = \frac{\sum_{k=1}^{n} (u_{ki})^m RSS_k}{\sum_{k=1}^{n} (u_{ki})^m} \end{cases} \quad (6)$$

The algorithm selected to classify RSS, KFCM, is featured by transforming RSS into Hilbert space, which is an infinite space. It is a dilemma that the infinite dimensional data cannot be stored. But in KFCM, there's no need to store the infinite dimensional data, and the key is calculating the distance in Hilbert space. In KFCM, the distance can be calculated by (7), $\Phi(x_k)$ and $\langle \cdot, \cdot \rangle$ stands for RSS in Hilbert space and the inner-multiple, respectively:

$$D_{ki} = \langle \Phi(RSS_k), \Phi(RSS_k) \rangle - \frac{2}{\sum_{k=1}^{n}(u_{ki})^m} \sum_{j=1}^{n} \langle \Phi(RSS_k), \Phi(RSS_j) \rangle$$
$$+ \frac{1}{\left( \sum_{k=1}^{n}(u_{ki})^m \right)^2} \sum_{j=1}^{n} \sum_{l=1}^{n} \langle \Phi(RSS_l), \Phi(RSS_j) \rangle$$
$$(7)$$

In this paper, Gaussian kernel functions are selected to compute the inner-multiple. And it calculates by (8), where $\lambda$ is the normalization factor:

$$\langle \Phi(RSS_k), \Phi(RSS_j) \rangle = e^{-\lambda \|RSS_j - RSS_k\|^2} \quad (8)$$

Via KFCM algorithm, RSS is labeled and then the classification information of RSS is used to matching RSS sampled by other terminals at different time.

### III. REALIZATION OF SEMI-SUPERVISED DISCRIMINANT EMBEDDING

After deploying the WILS and creating Radio Map, it is time to put forward the core algorithm in this paper. In this section, the theoretical analysis of SDE is presented and two optional alternatives of updating online are proposed.

#### A. Procedures of SDE

For convenience of demonstration, the steps of SDE algorithm are discussed first and then certify their details. Recalling the RSS that offline sampled Radio Map, except coordinates of RPs, $\{RSS_i\}_{i=1}^{m}$ are in $\Re^n$. And $m$ is determined by the number of RPs and $n$ equals to number of APs in the WILS. And each $RSS_i$ is labeled by using KFCM algorithm and its specific label is $c_i, i \in \{1, 2, \cdots, c\}$, where $c$ is the number of clusters. And all RSS are rewritten as $RSS_L = [RSS_1, RSS_2, \cdots, RSS_m] \in \Re^{n \times m}$. Then numerous of unlabeled RSS are got from online stage, such as RSS sampled by users who used the indoor position system. Let all unlabeled RSS symbols as $RSS_U = [RSS_1, RSS_2, \cdots, RSS_N] \in \Re^{n \times N}$. Then the proposed algorithm can be realized as following steps.

1. Construct neighbor graphs for labeled RSS. Let $G_L$ and $G'_L$ are two undirected graphs for all labeled RSS. In $G_L$, all nodes have the same class label and every node is in each node's KNN region. To $G'_L$, nodes has different class labels while every node is selected from KNN region of the node.

2. Label all the unlabeled RSS and update affinity graphs. By class matching, the label of RSS among $RSS_U$ is determined by two criterions stated by (4) and (9). In (9), $S_i$ denotes the similarity between RSS and cluster center. And $\text{cur}(\ )$ is a function to compute the curvature between two neighborhood RSS or the value of cluster centers. In (9), $\text{sgn}(\ )$ is a sign function to denote the number of slop variation less than $\varepsilon$. And it means the similarity between RSS and cluster center. The criterion stated in (9) is unlabeled $RSS_i$ belonged to class $c_i$ under the condition that $S_i$ is larger than the threshold of similarity $V_T$. If and only if a specific unlabeled RSS meets these two criterions, otherwise, the corresponding RSS would eliminate from $RSS_U$. After classifying all unlabeled RSS, a new set of unlabeled RSS is $RSS'_U \in \Re^{n \times k}, k < N$.

$$S_i = \text{sgn}(|\text{cur}(RSS_i) - \text{cur}(v_i)| \leq \varepsilon) \quad (9)$$

Combing $RSS_L$ and $RSS'_U$, a new kind of affinity graphs could be gained by step 1. But in the new graphs, $G$ and $G'$, include initially labeled RSS and unlabeled RSS. To distinguish these two kinds of RSS, the RSS in $RSS'_U$ would still denote as unlabeled although they have been classified by above two criterions.

3. Calculate affinity weights. Specify the affinity matrix $W$ of $G$ and $W'$ of $G'$, respectively. In (10) and (11), $w_{ij}$ and $w'_{ij}$ refer to the weight of the edge between $RSS_i$ and $RSS_j$ in $G$ and $G'$, respectively. And they can be calculated by (10) and (11) respectively. In (10), $t$ is normalization parameter and $M$ equals to the total number of RSS. Then the weights matrix of initial labeled RSS could also be computed by (10) and (11) respectively, and they are denoted as $W_L$ and $W'_L$ respectively.

$$W_{M \times M} = \begin{cases} w_{ij}^{ll} = \begin{cases} \exp(-\|RSS_i - RSS_j\|^2/t), & \text{if } RSS_i, RSS_j \in G; RSS_i \in RSS_L, RSS_j \in RSS_L \\ 0, & \text{others} \end{cases} \\ w_{ij}^{lu} = \begin{cases} \exp(-\|RSS_i - RSS_j\|^2/t), & \text{if } RSS_i, RSS_j \in G; RSS_i \in RSS_L, RSS_j \in RSS_U \\ 0, & \text{others} \end{cases} \\ w_{ij}^{uu} = \begin{cases} \exp(-\|RSS_i - RSS_j\|^2/t), & \text{if } RSS_i, RSS_j \in G; RSS_i \in RSS_U, RSS_j \in RSS_U \\ 0, & \text{others} \end{cases} \end{cases} \quad (10)$$

$$W'_{M \times M} = \begin{cases} w_{ij}'^{ll} = \begin{cases} \exp(-\|RSS_i - RSS_j\|^2/t), & \text{if } RSS_i, RSS_j \in G'; RSS_i \in RSS_L, RSS_j \in RSS_L \\ 0, & \text{others} \end{cases} \\ w_{ij}'^{lu} = \begin{cases} \exp(-\|RSS_i - RSS_j\|^2/t), & \text{if } RSS_i, RSS_j \in G'; RSS_i \in RSS_L, RSS_j \in RSS_U' \\ 0, & \text{others} \end{cases} \\ w_{ij}'^{uu} = \begin{cases} \exp(-\|RSS_i - RSS_j\|^2/t), & \text{if } RSS_i, RSS_j \in G'; RSS_i \in RSS_U', RSS_j \in RSS_U' \\ 0, & \text{others} \end{cases} \end{cases} \quad (11)$$

4. Compute dimensionality reduced OLD(DROLD). Find out the embedding matrix to map labeled $RSS_L$ in to intrinsic dimensionality which could be calculated by eigen-value algorithm proposed in [22]. And then the dimensionality reduced online localization database is got and symbolized as $RSS_{dr}$. The mentioned embedding matrix is computed by (12):

$$RSS_L(D_L' - W_L')RSS_L^T M = \lambda RSS_L(D_L - W_L)RSS_L^T M \quad (12)$$

Where $D$ and $D'$ are diagonal matrices with diagonal elements $d_{ii} = \sum_{j=1}^{m} w_{ij}$ and $d_{ii}' = \sum_{j=1}^{m} w_{ij}'$. In (12), M is the embedding matrix and $RSS_{dr}$ is calculated by (13):

$$RSS_{dr} = M^T \cdot RSS_L \quad (13)$$

5. Compute online transformation eigen-matrix. After adding new RSS into $RSS_L$, a refreshed online position eigen-matrix can be computed by an equation which is the same with (12). But the affinity matrices and inputting RSS are the updated ones.

### B. Details of SDE and alternatives of updating RSS

In the following paragraphs, detailed justifications for the above realization of SDE are proposed. In SDE algorithm, the core features are updating Radio Map with unlabeled RSS and reducing dimension of labeled RSS as DROLD as well as computing eigen transforming matrix by the amalgamate RSS.

The goal of SDE is the same with LDE that proposed in [20] and its optimization problem can be described by (14). But in (14), there is no specification to point out which set of RSS and affinity are selected to use in the equation. And the analyzing processes of SDE are similar to LDE. Only one difference between these two algorithms is that the affinity weight matrix of amalgamated RSS is a combinatorial matrix. Therefore, the theoretical analysis of SDE can be expressed the same as LDE.

$$\max \quad J(v) = \sum_{i,j} \|v^T RSS_i - v^T RSS_j\| w_{ij}'$$
$$s.t. \quad \sum_{i,j} \|v^T RSS_i - v^T RSS_j\| w_{ij} = 1 \quad (14)$$

Above analysis about the details of SDE is the first key feature of SDE. And another feature of SDE is the updating of RSS. But there are options of refreshing RSS. In this paper, alternatives are proposed corresponding to two position stages in WILSs. First option is updating OLD offline. It means that RSS used to update OLD comes from the RSS sampled by the users who used the specified WILS. As to the way to transmit the sampled RSS to other users is elide in this paper and it might be a good idea for future work.

Another method of refreshing RSS is according to the RSS sampled by specified user during online stage. Then RSS which is satisfied to the two criterions is added into the OLD one by one.

### C. Flowchart of SDE applied in WLPS

In Fig.3, it shows the process of applying SDE in KNN algorithm. But there is alternatives to compute the eigen matrix as it is presented in above paragraphs. In this flowchart, there is no special distinction about these two optional methods of refreshing RSS. In the next section, the simulation would demonstrate the influence of the number of RSS updated. And it can be treated as an example to show the efficiency of online refreshing.

During offline phase, Radio Map is constructed by using a software, Netstumbler, installed on a Lenovo laptop(V450) whose OS is Windows XP to sample RSS available at specified RP. And then an intrinsic dimensionality estimation algorithm is used to compute the intrinsic dimensionality of constructed Radio Map. Both of Radio Map and its intrinsic dimensionality are the input data of the proposed algorithm. By using CM-SDE algorithm, DROLD and eigen matrix are computed.

A RSS is received by mobile terminals or laptops multiples with eigen matrix that is computed offline. Then the dimensionality reduced online RSS could match with DROLD by KNN to calculate the location.

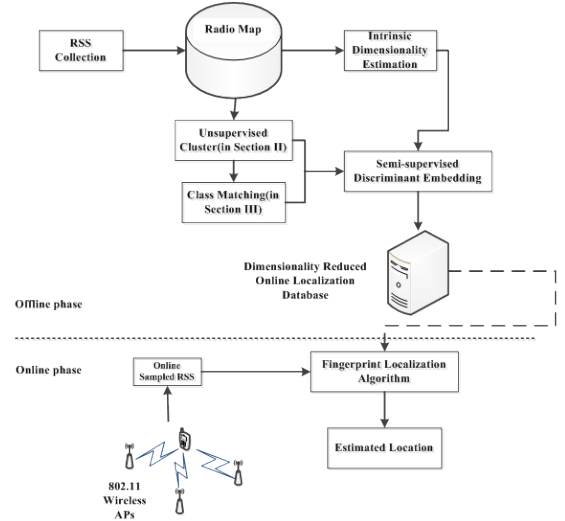

Fig.3: Flowchart of CM-SDE applied in WILS. Radio Map is created at offline stage and then an intrinsic dimensionality estimation algorithm is used to estimate the intrinsic dimensionality of it. Both of them are the input of SDE algorithm. The work of offline stage finished until the output of DROLD and eigen matrix are got. During online stage, a sample RSS multiples with eigen matrix is calculated at offline stage to compute the reduced dimensionality of online sampled RSS. Then the online localization output is computed by KNN algorithm.

## IV. NUMERICAL ANALYSIS

### A. Settings of Test Beds

Simulations to examine the performance of the proposed SDE algorithm for WILSs are according to the testing data obtained from an office environment. And a series of experiments are carried out in an office environment that mimics the features of a typical WILS.

The experiments were carried out on the 12th floor of 2A building in scientific park of Harbin Institute of Technology as was shown in Fig.1. The test area consists of one hallway and 21 rooms and the main tests are in the hallway. By dividing the whole test area into several $0.5 \times 0.5$ grids, there are 1000 RPs in total and all RPs are in the hallway. All the RSS measurements were recorded by software, named Netstumbler, installed on a Lenovo laptop (V450) with Windows XP operation system. There are 27 APs deployed in the whole interested area, which is a relative AP-rich area. At each RP, 100 samples from four directions(east, south, west and north) are sampled and it means that every RP sampled 400 samples. As the method that used to divide subareas stated in [23], a similar scheme is applied in this paper. The interested localization zone is clustered into six sub-areas, and denoted as A1, A2, A3, A4, A5 A6, respectively. In the following performance analysis of the proposed algorithms, the main idea would focus on analyzing parameters and localization performance in A1 area. The simulating consequences of other areas are presented as appendix.

### B. Optimization of parameters

The proposed algorithm in this paper has several input parameters. The input data include the initial labeled RSS and the unlabeled RSS. And there are another four parameters to determine the performance of dimensionality reduction. They are intrinsic dimension of RSS, number of clusters, ranges of affinity and normalized factor. In the proposed algorithm, all the four parameters above only can be integer, therefore, a sequence of discrete analyzes on each parameter, and then the partial optimization of parameter could be gained. During analyzing parameters, the performance of localization accuracy is selected to show cons-pros of specified value of the parameter.

*1) Analysis of Intrinsic Dimensionality*

The first parameter optimized is the intrinsic dimensionality. Although the intrinsic dimensionality estimation algorithm, like eigen-value, could compute the approximate dimensionality of RSS, it is just a kind of roughly estimation. And this is one of the most significant parameters which determine the accuracy of dimensionality reduction as well as the calculation of transforming matrix. Therefore, the first step is going to find out or to demonstrate its influence on localization accuracy as the increasing value of intrinsic dimensionality. And during simulating this parameter, other parameters are set as default value.

As it is shown in Fig.4, there are 9 specified results and the results of KNN algorithm. Comparing to KNN algorithm, DROLD could improve the accuracy of positioning. Taking error radius equals to 1 meter as an example, the maximum improvements of localization accuracy is up to 5 percent comparing to KNN algorithm. Through above figure, the local optimal value of intrinsic dimensionality could be 4, 5, or 6. These three values would be taken as preferred value of intrinsic dimensionality in A1 zone when carrying following parameter simulations.

*2) Analysis of number of clusters*

Taking above local optimal of intrinsic dimensionality as prerequisite, a local optimization of the number of clusters will be presented in this part. As the description in Section III, the classification label of RSS determines the construction of affinity graphs. Therefore, the parameter that is optimized in this part is the number of clusters.

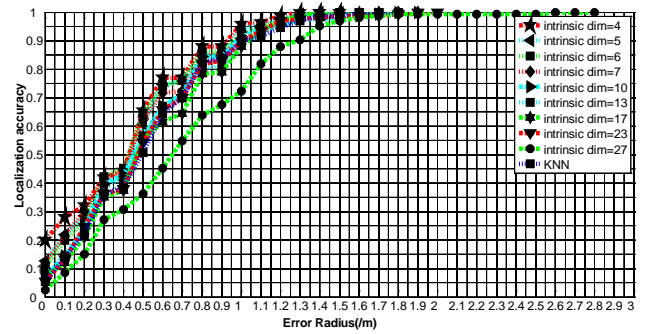

Fig.4: Results of simulating intrinsic dimensionality in A1 area. In the deployed localization system, there are 27 APs. In the initial simulation, the value of intrinsic dimensionality increases from 1 to 27 one by one. But to show the results clearly, there are 9 specified results and the result of KNN algorithm.

As it shows in Fig.5, the localization accuracy does not change to the specified error radius while the number of clusters is changing. And the value of intrinsic dimensionality equals to 4 during simulating the influence of changing the number of clusters. According to the outputs, it is a good choice to let the number of clusters equals to 2 for it could optimize the localization accuracy locally as well as reduce the computation complexity of the algorithm.

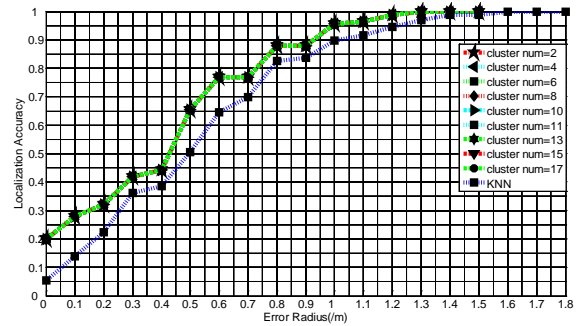

Fig.5: Output of the value intrinsic dimensionality equals to 4 as the number of clusters ranges in positive integers. As it shows in the figure, the changes of number of clusters in A1 zone do not have influence on the performance of SDE's efficiency applying in WILS and it is one of the reasons that there are only 9 specified values of the number of cluster plotted in above figure. Therefore, in order to reduce the complexity of algorithm, the value of clusters equals to the minimal feasible optimal value is shown in above figure.

*3) Analysis of ranges of affinity*

Next parameter optimized in this part is the range of affinity. It means to optimize the range of neighborhood of RSS. There are two prevalent definitions of ranges of affinity: KNN and $\varepsilon$ ball. In this paper, KNN is a preferred method and in following simulation focuses on optimizing the number of neighbors.

In Fig.6, the results are almost the same with Fig.5. There is an explanation to these two figures for the localization accuracy has little relationship with these two parameters in A1 area. Let the value of range of affinity equals to 6 in order to reduce the computation complexity.

*4) Analysis of normalized factor and updating ratio*

To show the influence of normalized factor, the detailed theoretical analysis of it could be found in [24]. As from the simplified analysis, the value of normalized factor should have no influence on the localization accuracy for normalized factor is just a positive value to ensure the optimization process could be solved by eigen decomposition. In this part, the simulation is going to present that the value of normalized factor should be as small as possible under the condition that the objection could be calculated by eigen-decomposition.

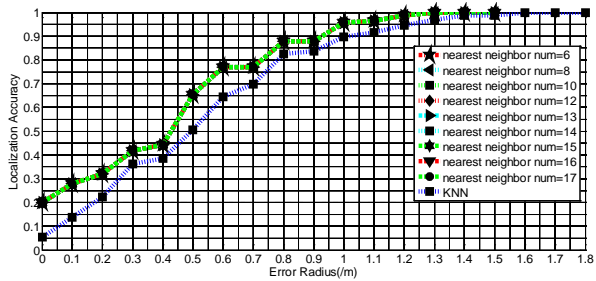

Fig.6: Results of simulating ranges of affinity as the intrinsic value is 4 and the number of clusters is 2 in A1 area. Comparing to Fig.5, the outputs of them are almost the same with each other. There is an explanation to these two figures for the localization accuracy has little relationship to these two parameters in A1 area. Therefore, as the choice of the number of clusters, the value of ranges of affinity also equals to the minimal feasible value.

In Fig.7, it shows a stabilization process as the value of normalized factor decreased to $10^{-8}$. And it clearly shows that as the normalized factor decreased to the feasible value, the localization accuracy is becoming more stable.

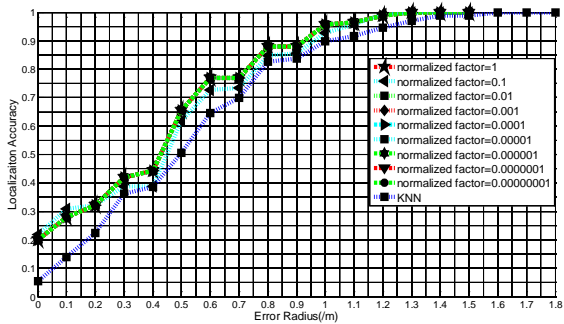

Fig.7: The influence of normalized factor in A1. As it shows in above figure, the minimal value that could maintain semi-positive definite of eigen-decomposition is $10^{-8}$ in A1 area. And it is clear that the proposed algorithm becomes more stable as the decreasing of normalized factor.

And in this part, there is another parameter to be analyzed, the refreshing ratio. This is a key parameter to show the influence on localization accuracy as the changes of ratio on updating the RSS. For the limitation of number of sampled RSS, the value ratio ranges from one half to 10 percent.

The follow figure would provide a good scheme to select proper ratio to update RSS. As it shows in Fig.8, there is a trend that with the increasing number of refreshed RSS, the localization accuracy increased as well. But it does not mean that infinite increasing the number of updated RSS could infinite increasing the localization accuracy. As it shows in the figure, the optimal ration is about one fifths in A1 zone.

## C. Performance Analysis

Through the analyzing on each parameters of CM-SDE, the local optimized parameters of A1 zone could be obtained. In this part, comparisons between LDE based localization approach joint with KNN(LDE-KNN) and the proposed algorithm(SDE-KNN) are reported. And there is also a comparison to the initial KNN localization method. The setting parameteres of SDE-KNN are according to above analysis. And the parameteres of LDE-KNN are the same with SDE-KNN except the parameter of ratio for there is no updating of RSS in LDE-KNN algorithm.

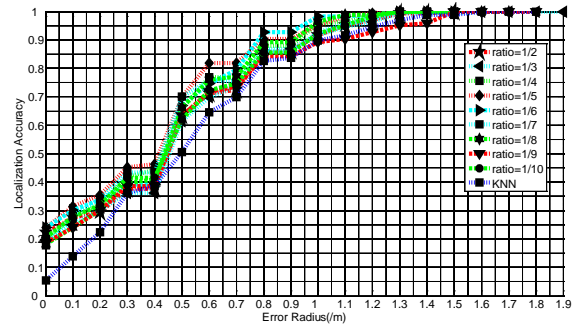

Fig.8: As it shows in above figure, the blue line means the ratio equals to 1. It means that there is no RSS updated. As the increasing the number of updating RSS, the localization accuracy is increasing. But it is not means that infinite increasing the updated RSS could be infinite increasing the localization accuracy. As it shows in the figure, there is also an optimal value of ratio. In A1 zone, the optimal of ratio might be one fifths.

After optimizing the parameters of SDE-KNN, the performance of it is overly better than initial KNN and LDE-KNN. In Fig.9, it clearly presents that localization accuracy of SDE-KNN is higher than 70% as the error radius is 0.5 meter. And it is about 10 percent and 25 percent higher than KNN and LDE-KNN respectively. As the error radius equals to 1 meter, localization accuracy of the proposed algorithm is up to 95% in the WILS.

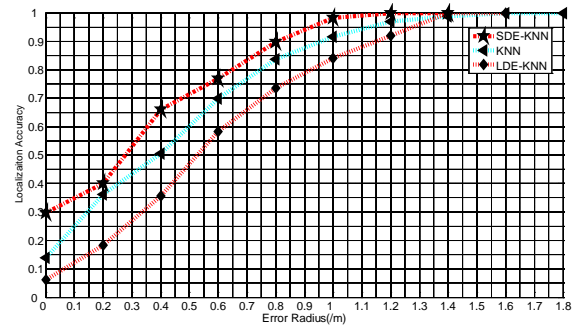

Fig.9: Performance comparison between SDE-KNN, KNN and LDE-KNN in A1 zone. As it shows in the figure, the proposed could improve the localization accuracy. And the performance of it is much better than LDE-KNN.

## V. CONCLUSION

In this paper, a semi-supervised RSS dimensionality reduction algorithm is proposed and there are lots of analyses about the theoretical realization of the proposed method. Another significant innovation of this paper is jointing the fingerprint based algorithm with CM-SDE algorithm to improve the localization accuracy of indoor position. Comparing with LDE-KNN algorithm, SDE-KNN method is going to update RSS during online stage or offline stage. It is the update scheme that improves the performance of the proposed algorithm.

After locally analyzing of parameters, the optimized value of each parameter could be gained. As it is presented in numeral analysis section, the performance optimized parameters of SDE-KNN is overly better than initial KNN and LDE-KNN. The localization accuracy of SDE-KNN is higher than 70% as the error radius is 0.5 meter. And it is about 10 percent and 25 percent higher than KNN and LDE-KNN respectively. As the error radius equals to 1 meter, localization accuracy of the proposed algorithm is up to 95% in the WILS.

However, there is a few of undone works in this paper. The first one is the global optimization of SDE algorithm. There are four different parameters needed to optimize at the same if applying global optimization. Another left work of this paper is the fast construction of Radio Map. In this paper, there is no real time Radio Map during the experiments for the construction of Radio Map is a time-consuming process.

## APPENDIX

In this appendix, the simulating results of other five sub-areas, denoted as A2 to A6, are demonstrated. Since the value of normalized factor has little effect on the localization accuracy, therefore, there is no result about the analysis of normalized factor in the following presentation. From Fig.10 to Fig.14 presents the results of simulating on parameters and overall localization accuracy performance of other five sub-areas. In each figure, there are five figures to show the results of analysis of intrinsic dimensionality, analysis of number of clusters, analysis of ranges of affinity, analysis of updating ratio and overall performance analysis, respectively.

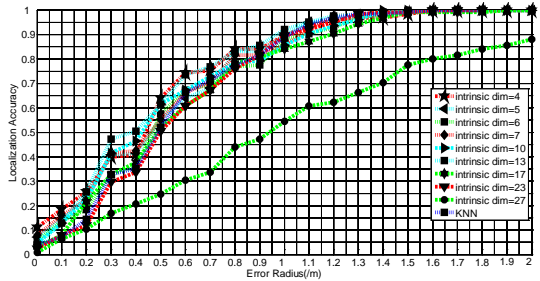

(1) Analysis of intrinsic dimensionality in A2

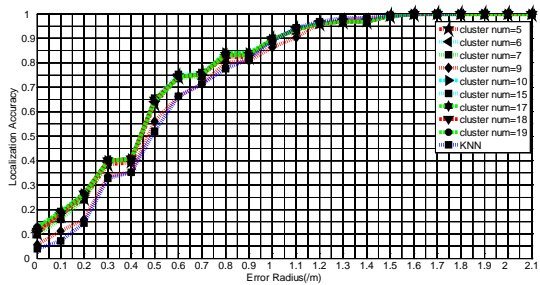

(2) Analysis of number of clusters in A2 area

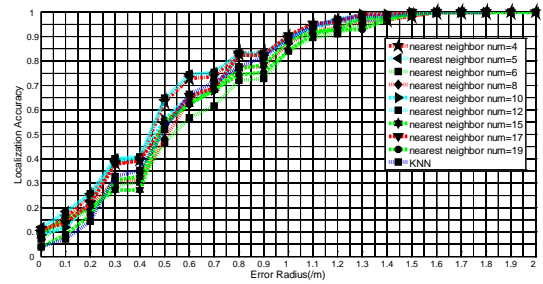

(3) Analysis of ranges of affinity in A2 area

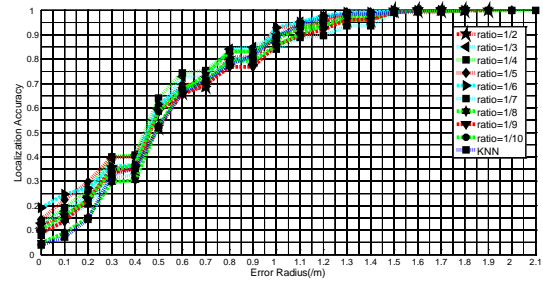

(4) Analysis of updating ratio in A2 area

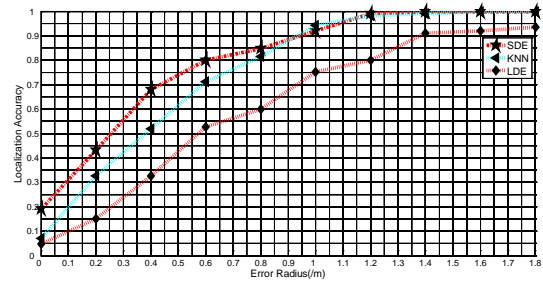

(5) Performance Analysis in A2 area

Fig.10: Above five figures present the results of parameter analysis and performance analysis in A2 area. From (1) to (5), it presents analysis of intrinsic dimensionality, analysis of number of clusters, analysis of ranges of affinity, analysis of updating ratio and overall performance analysis, respectively.

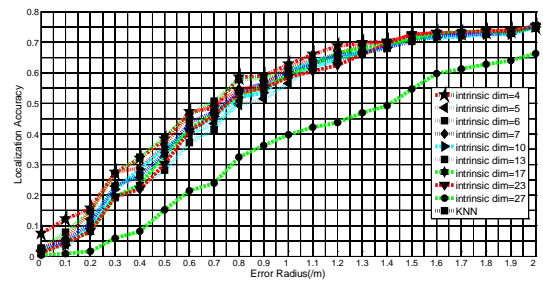

(1) Analysis of intrinsic dimensionality in A3 area

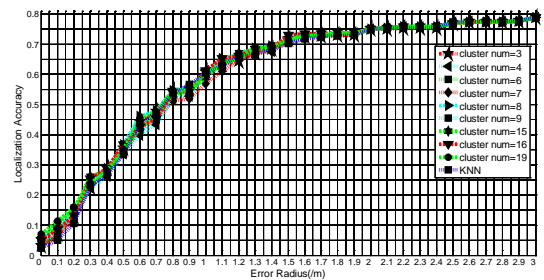

(2) Analysis of number of clusters in A3 area

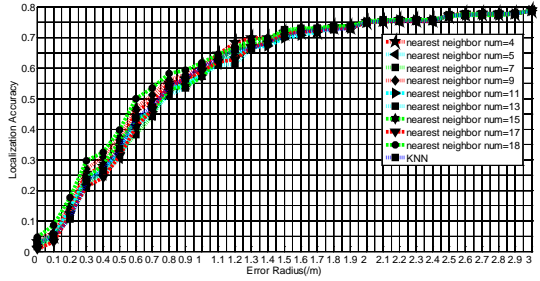

(3) Analysis of ranges of affinity in A2 area

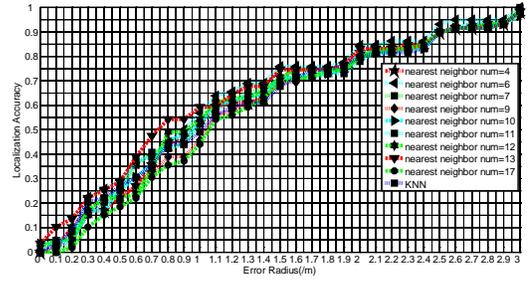

(3) Analysis of ranges of affinity in A4 area

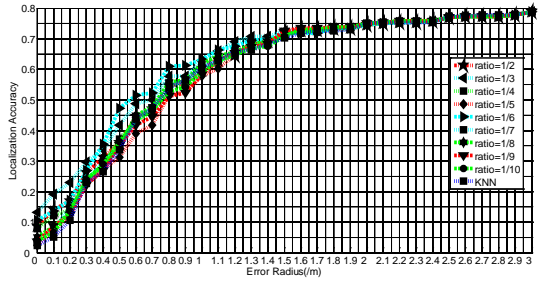

(4) Analysis of updating ratio in A3 area

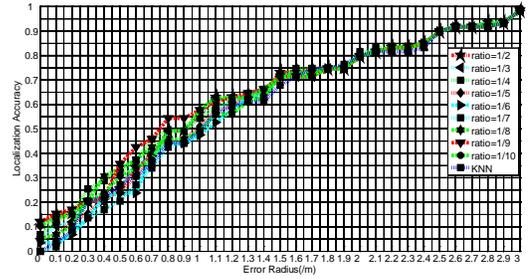

(4) Analysis of updating ratio in A4 area

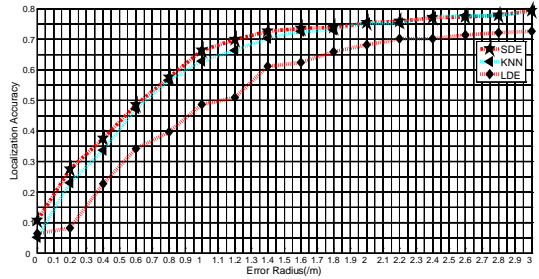

(5) Performance Analysis in A3 area

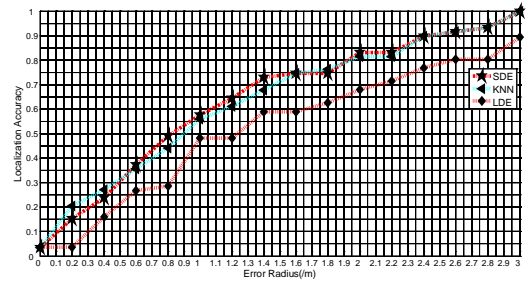

(5) Performance Analysis in A4 area

Fig.11: Above five figures present the results of parameter analysis and performance analysis in A3 area. From (1) to (5), it presents analysis of intrinsic dimensionality, analysis of number of clusters, analysis of ranges of affinity, analysis of updating ratio and overall performance analysis, respectively.

Fig.12: Above five figures present the results of parameter analysis and performance analysis in A4 area. From (1) to (5), it presents analysis of intrinsic dimensionality, analysis of number of clusters, analysis of ranges of affinity, analysis of updating ratio and overall performance analysis, respectively.

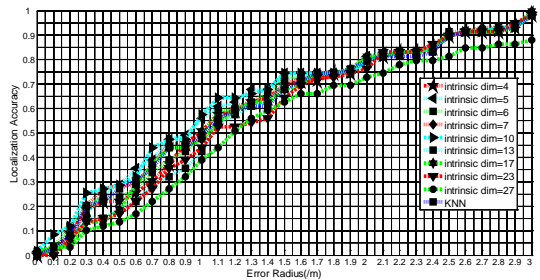

(1) Analysis of intrinsic dimensionality in A4 area

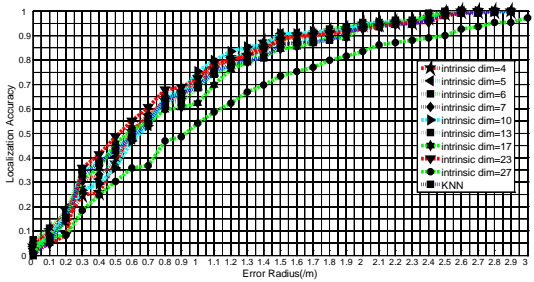

(1) Analysis of intrinsic dimensionality in A5 area

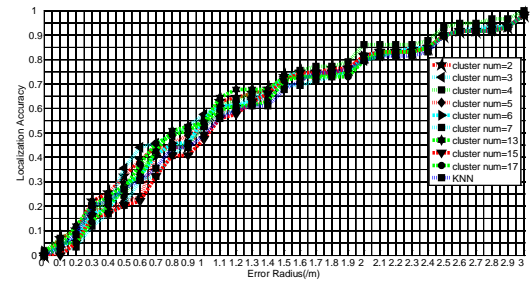

(2) Analysis of number of clusters in A4 area

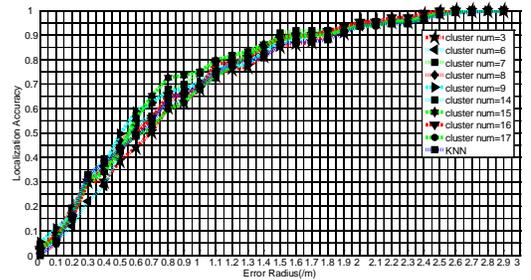

(2) Analysis of number of clusters in A5 area

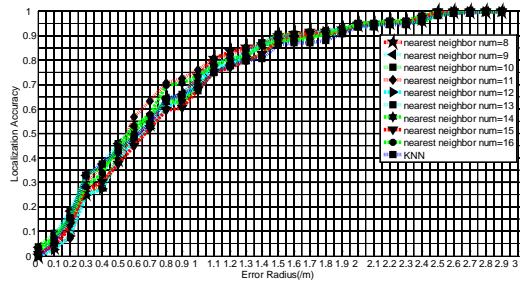

(3) Analysis of ranges of affinity in A5 area

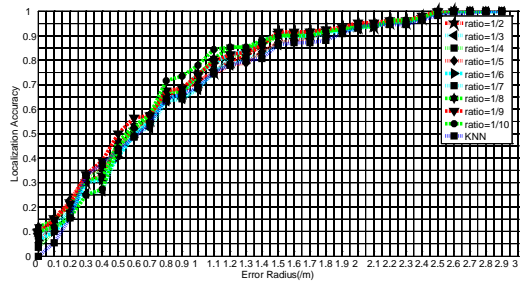

(4) Analysis of updating ratio in A5 area

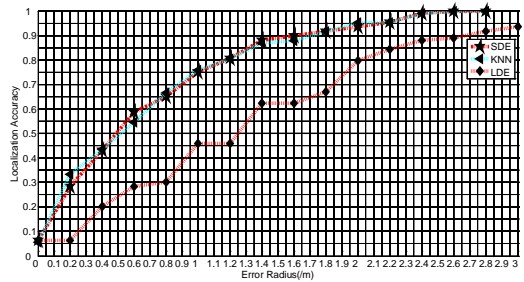

(5) Performance Analysis in A5

Fig.13: Above five figures present the results of parameter analysis and performance analysis in A5 area. From (1) to (5), it presents analysis of intrinsic dimensionality, analysis of number of clusters, analysis of ranges of affinity, analysis of updating ratio and overall performance analysis, respectively.

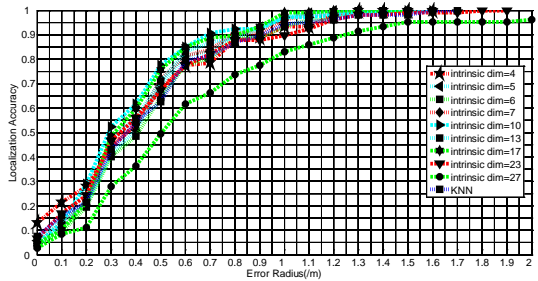

(1) Analysis of intrinsic dimensionality in A6 area

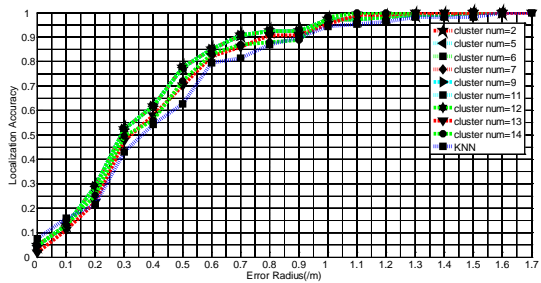

(2) Analysis of number of clusters in A6 area

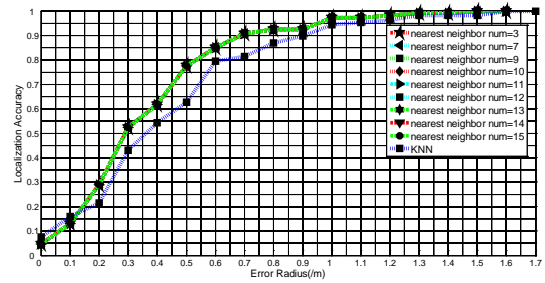

(3) Analysis of ranges of affinity in A6 area

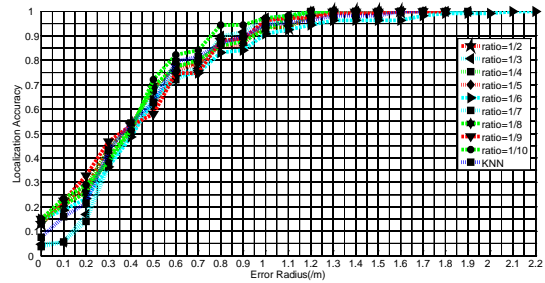

(4) Analysis of updating ratio in A6 area

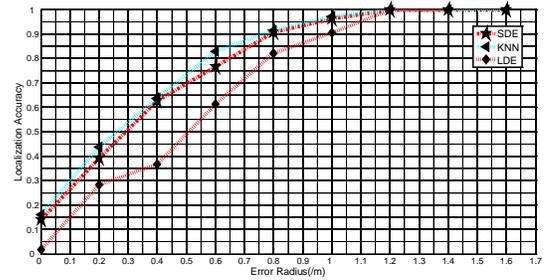

(5) Performance Analysis in A6 area

Fig.14: Above five figures present the results of parameter analysis and performance analysis in A6 area. From (1) to (5), it presents analysis of intrinsic dimensionality, analysis of number of clusters, analysis of ranges of affinity, analysis of updating ratio and overall performance analysis, respectively.

ACKNOWLEDGMENTS

This work was supported in part by National Natural Science Foundation of China (Grant No. 61101122 and 61071105), Postdoctoral Science-Research Development Foundation of Heilongjiang Province (Grant No. LBH-Q12080) and Specialized Research Fund for the Doctoral Program of Higher Education (Grant No. 20122301120004). The authors thank a lot for the group members, who contribute hard work and valuable suggestions.

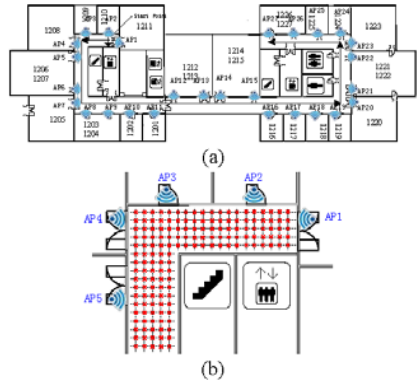

Fig.1: Deployment of WILS in the experimental zone and its sampling grid which interval equals 0.5 meter. (a): In the WLAN-based indoor, there are 27 APs and each AP works on 2.4GHz, which is Industrial Scientific Medical(ISM) band and is freedom of license. And the test zone is 12th floor of 2A building in scientific park of Harbin Institute of Technology (HIT). The interested area of indoor localization is the hallway. (b): By enlarging part of the test zone, the details of RPs, marked by red spot, are showed in it. In this paper, the interval of reference point is 0.5 meter.

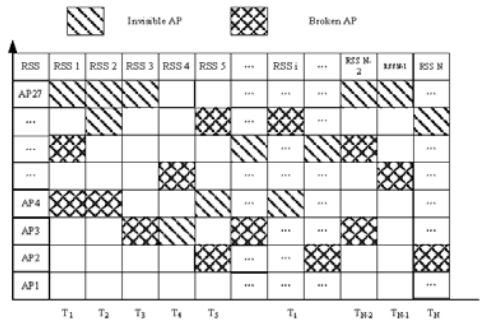

Fig.2: At a specified reference point, RSS is sampled by the mobile terminals or laptops. And the sampled RSS is treated as the discrete time-sequence. During the process of sampling RSS, there are several APs are invisible or broken down, especially during online stage for it could be ensure that all APs are opening and working on when constructing Radio Map.

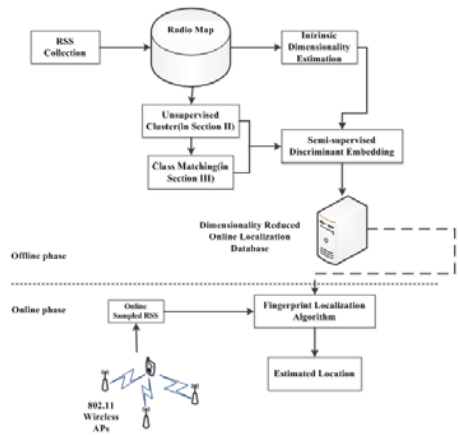

Fig. 3: Flowchart of CM-SDE applied in WILS. Radio Map is created at offline stage and then an intrinsic dimensionality estimation algorithm is used to estimate the intrinsic dimensionality of it. Both of them are the input of SDE algorithm. The work of offline stage finished until the output of DROLD and eigen matrix are got. During online stage, a sample RSS multiples with eigen matrix is calculated at offline stage to compute the reduced dimensionality of online sampled RSS. Then the online localization output is computed by KNN algorithm.

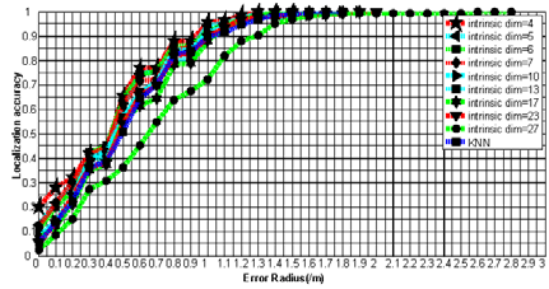

Fig.4: Results of simulating intrinsic dimensionality in A1 area. In the deployed localization system, there are 27 APs. In the initial simulation, the value of intrinsic dimensionality increases from 1 to 27 one by one. But to show the results clearly, there are 9 specified results and the result of KNN algorithm

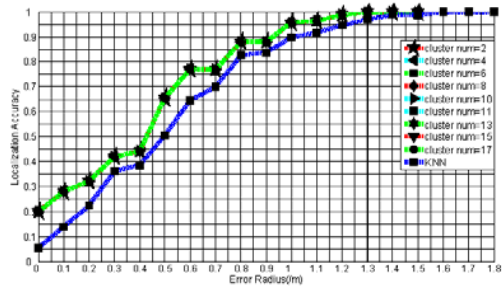

Fig.5: Output of the value intrinsic dimensionality equals to 4 as the number of clusters ranges in positive integers. As it shows in the figure, the changes of number of clusters in A1 zone do not have influence on the performance of SDE's efficiency applying in WILS and it is one of the reasons that there are only 9 specified values of the number of cluster plotted in above figure. Therefore, in order to reduce the complexity of algorithm, the value of clusters equals to the minimal feasible optimal value is shown in above figure.

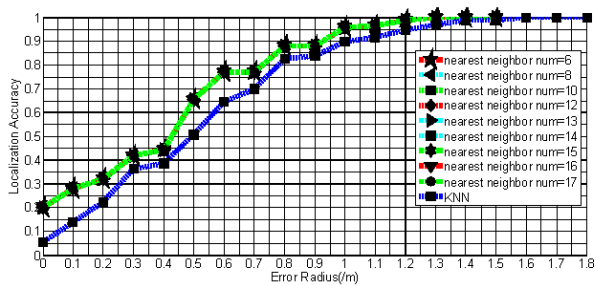

Fig.6: Results of simulating ranges of affinity as the intrinsic value is 4 and the number of clusters is 2 in A1 area. Comparing to Fig.5, the outputs of them are almost the same with each other. There is an explanation to these two figures for the localization accuracy has little relationship to these two parameters in A1 area. Therefore, as the choice of the number of clusters, the value of ranges of affinity also equals to the minimal feasible value.

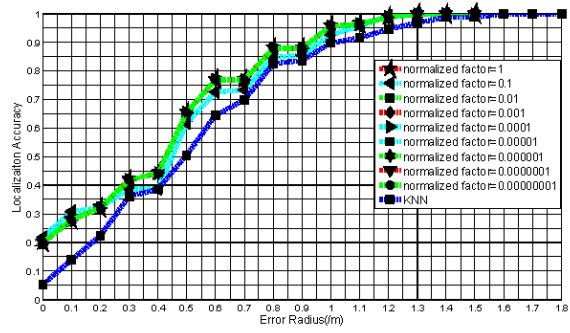

Fig.7: The influence of normalized factor in A1. As it shows in above figure, the minimal value that could maintain semi-positive definite of eigen-decomposition is $10^{-8}$ in A1 area. And it is clear that the proposed algorithm becomes more stable as the decreasing of normalized factor.

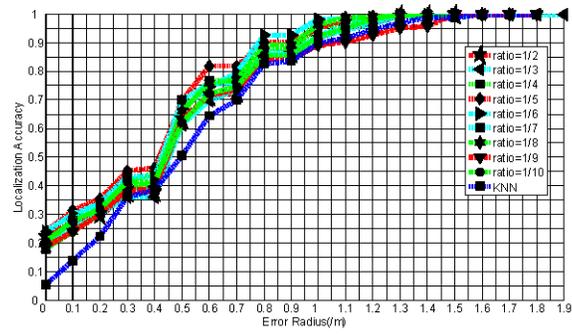

Fig.8: As it shows in above figure, the blue line means the ratio equals to 1. It means that there is no RSS updated. As the increasing the number of updating RSS, the localization accuracy is increasing. But it is not means that infinite increasing the updated RSS could be infinite increasing the localization accuracy. As it shows in the figure, there is also an optimal value of ratio. In A1 zone, the optimal of ratio might be one fifths.

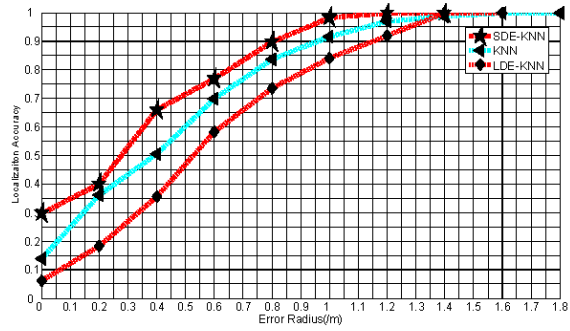

Fig.9: Performance comparison between SDE-KNN, KNN and LDE-KNN in A1 zone. As it shows in the figure, the proposed could improve the localization accuracy. And the performance of it is much better than LDE-KNN.

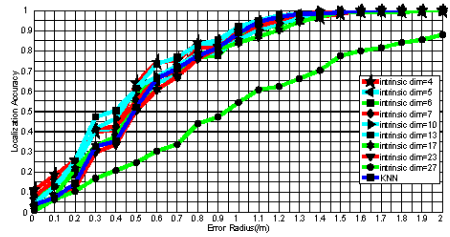

(1) Analysis of intrinsic dimensionality in A2

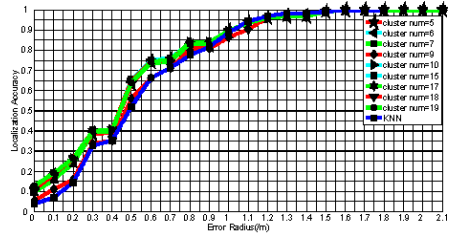

(2) Analysis of number of clusters in A2 area

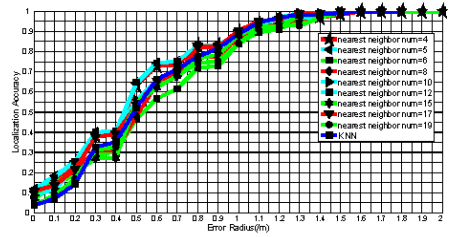

(3) Analysis of ranges of affinity in A2 area

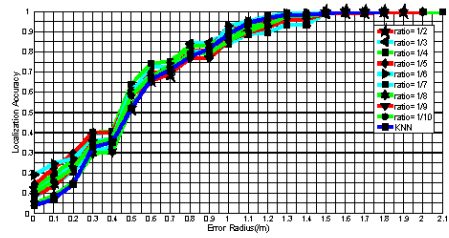

(4) Analysis of updating ratio in A2 area

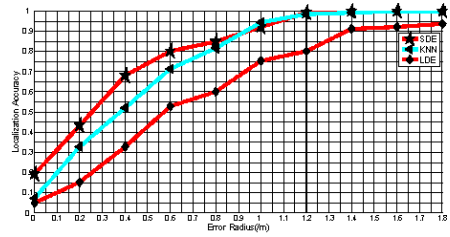

(5) Performance Analysis in A2 area

Fig.10: Above five figures present the results of parameter analysis and performance analysis in A2 area. From (1) to (5), it presents analysis of intrinsic dimensionality, analysis of number of clusters, analysis of ranges of affinity, analysis of updating ratio and overall performance analysis, respectively.

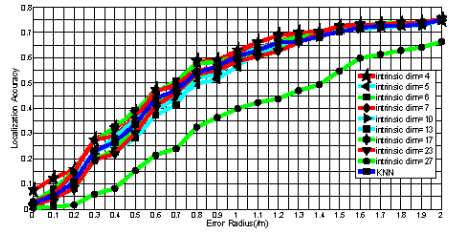

(1) Analysis of intrinsic dimensionality in A3 area

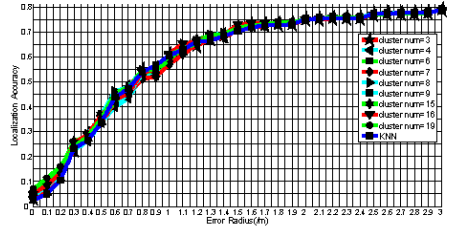

(2) Analysis of number of clusters in A3 area

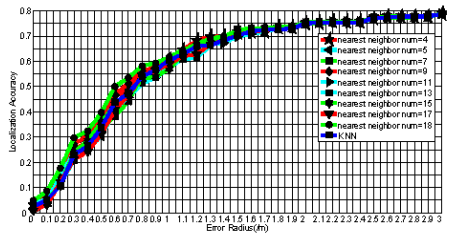

(3) Analysis of ranges of affinity in A2 area

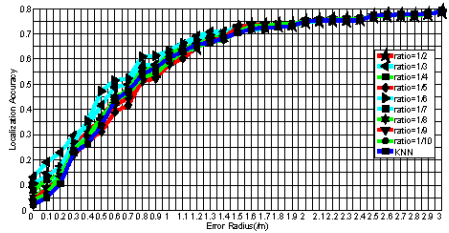

(4) Analysis of updating ratio in A3 area

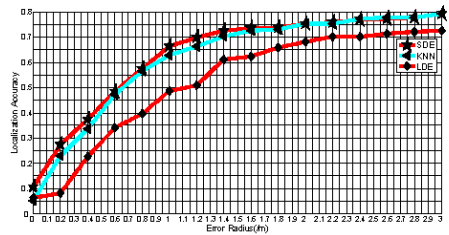

(5) Performance Analysis in A3 area

Fig.11: Above five figures present the results of parameter analysis and performance analysis in A3 area. From (1) to (5), it presents analysis of intrinsic dimensionality, analysis of number of clusters, analysis of ranges of affinity, analysis of updating ratio and overall performance analysis, respectively.

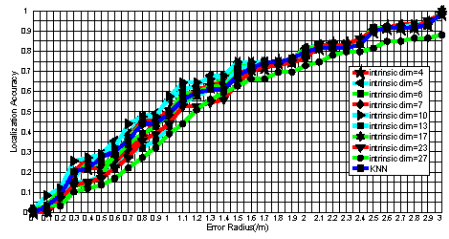

(1) Analysis of intrinsic dimensionality in A4 area

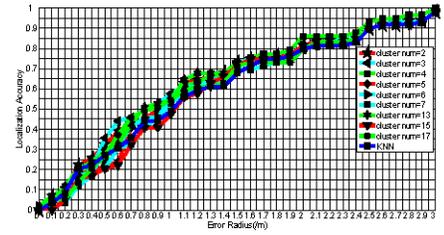

(2) Analysis of number of clusters in A4 area

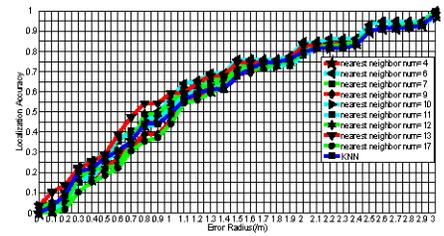

(3) Analysis of ranges of affinity in A4 area

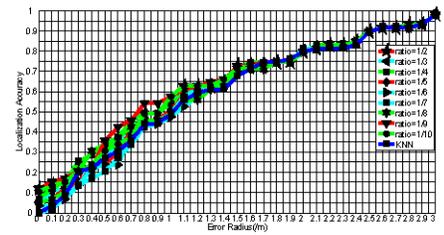

(4) Analysis of updating ratio in A4 area

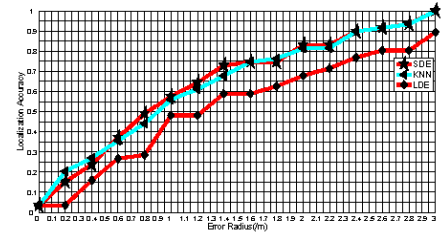

(5) Performance Analysis in A4 area

Fig.12: Above five figures present the results of parameter analysis and performance analysis in A4 area. From (1) to (5), it presents analysis of intrinsic dimensionality, analysis of number of clusters, analysis of ranges of affinity, analysis of updating ratio and overall performance analysis, respectively.

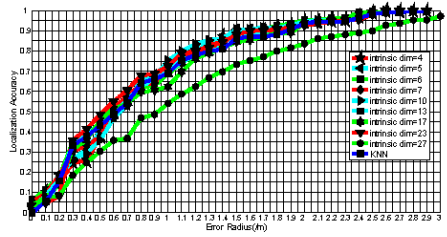

(1) Analysis of intrinsic dimensionality in A5 area

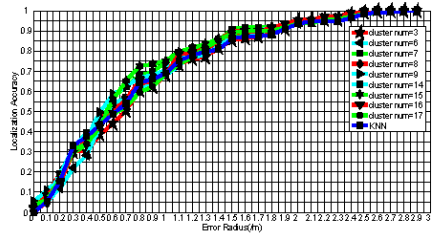

(2) Analysis of number of clusters in A5 area

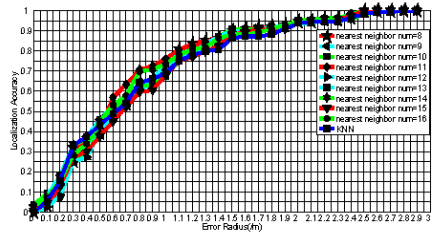

(3) Analysis of ranges of affinity in A5 area

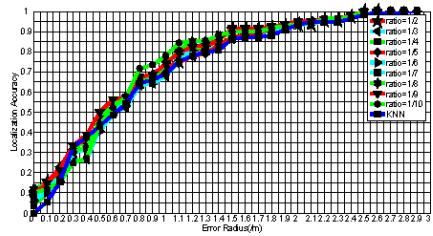

(4) Analysis of updating ratio in A5 area

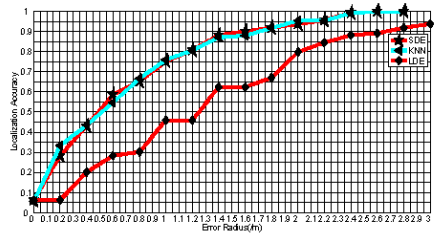

(5) Performance Analysis in A5

Fig.13: Above five figures present the results of parameter analysis and performance analysis in A5 area. From (1) to (5), it presents analysis of intrinsic dimensionality, analysis of number of clusters, analysis of ranges of affinity, analysis of updating ratio and overall performance analysis, respectively.

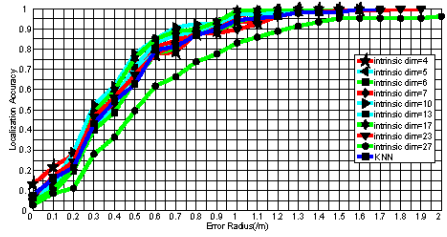

(1) Analysis of intrinsic dimensionality in A6 area

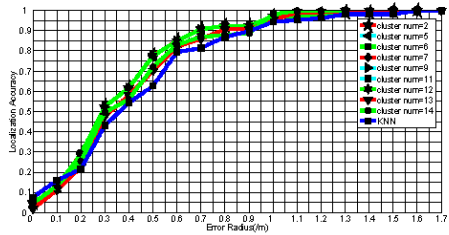

(2) Analysis of number of clusters in A6 area

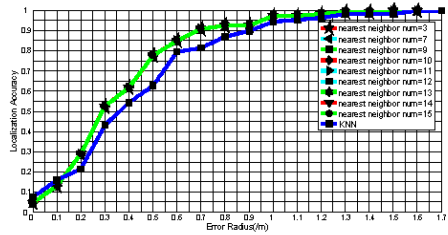

(3) Analysis of ranges of affinity in A6 area

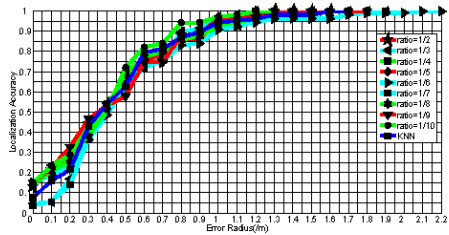

(4) Analysis of updating ratio in A6 area

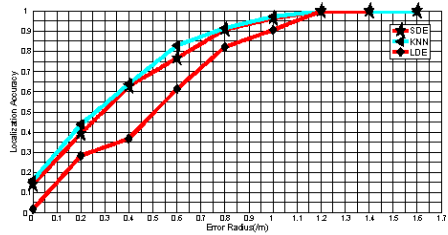

(5) Performance Analysis in A6 area

Fig.14: Above five figures present the results of parameter analysis and performance analysis in A6 area. From (1) to (5), it presents analysis of intrinsic dimensionality, analysis of number of clusters, analysis of ranges of affinity, analysis of updating ratio and overall performance analysis, respectively.